\documentclass[runningheads]{llncs}
\usepackage[T1]{fontenc}
\usepackage{url}
\usepackage{booktabs}
\usepackage{subcaption} 
\usepackage{graphicx}
\usepackage{multirow}
\usepackage{listings}
\begin{document}
\title{VLN-Pilot: Large Vision-Language Model as an Autonomous Indoor Drone Operator}
%
%
\author{
Bessie Dominguez-Dager*\orcidID{0000-0003-2214-3849} \and
Sergio Suescun-Ferrandiz\orcidID{0009-0008-1521-0031} \and
Felix Escalona\orcidID{0000-0003-2245-601X} \and
Francisco Gomez-Donoso\orcidID{0000-0002-7830-2661} \and
Miguel Cazorla\orcidID{0000-0001-6805-3633}
}
\authorrunning{B. Dominguez-Dager et al.}
\titlerunning{VLN-Pilot}
%
\institute{University Institute for Compute Research. University of Alicante. \\
Ctra. San Vicente del Raspeig SN, 03690, Alicante. SPAIN. \\
\email{\{bessie.dominguez, sergio.suescun, felix.escalona, fgomez, miguel.cazorla\}@ua.es}\\
{*}Corresponding Author}
\maketitle              

\begin{abstract}
This paper introduces \textbf{VLN-Pilot}, a novel framework in which a large Vision-and-Language Model (VLLM) assumes the role of a human pilot for indoor drone navigation. By leveraging the multimodal reasoning abilities of VLLMs, VLN-Pilot interprets free-form natural language instructions and grounds them in visual observations to plan and execute drone trajectories in GPS-denied indoor environments. Unlike traditional rule-based or geometric path-planning approaches, our framework integrates language-driven semantic understanding with visual perception, enabling context-aware, high-level flight behaviors with minimal task-specific engineering. VLN-Pilot supports fully autonomous instruction-following for drones by reasoning about spatial relationships, obstacle avoidance, and dynamic reactivity to unforeseen events. We validate our framework on a custom photorealistic indoor simulation benchmark and demonstrate the ability of the VLLM-driven agent to achieve high success rates on complex instruction-following tasks, including long-horizon navigation with multiple semantic targets. Experimental results highlight the promise of replacing remote drone pilots with a language-guided autonomous agent, opening avenues for scalable, human-friendly control of indoor UAVs in tasks such as inspection, search-and-rescue, and facility monitoring. Our results suggest that VLLM-based pilots may dramatically reduce operator workload while improving safety and mission flexibility in constrained indoor environments.


\end{abstract}
\section{Introduction}
Vision-and-Language Navigation (VLN) has emerged as a promising paradigm for enabling embodied agents to interpret natural language instructions and navigate complex environments by grounding their actions in visual observations. While significant advances have been made in ground-based robotic navigation, extending VLN to autonomous drones remains an open and challenging frontier. Drones operating indoors face unique difficulties, such as constrained spaces, dynamic obstacles, and the absence of GPS signals, making human-in-the-loop piloting still the standard practice in many industrial applications. \\

In this work, we investigate a novel framework in which large Vision-Language Models (VLLMs), including state-of-the-art conversational agents such as GPT and Gemini, assume the role traditionally filled by a remote human pilot. Our system aims to replace or augment the pilot by leveraging the reasoning abilities of these models to interpret visual inputs and make high-level control decisions. The drone is provided only with a topological map of the indoor environment, describing the room connectivity graph, along with frontal images captured during flight. In this constrained sensory setting, the VLLM must reason about its current location, interpret language-driven mission goals, and plan semantic trajectories. \\

For low-level control, we employ a state-machine-based architecture that manages stable flight, obstacle avoidance, and basic drone commands. The VLLM further acts as a supervisory controller, deciding when to transition between states based on its understanding of the navigation objective and the visual scene. This hybrid design enables interpretable, robust, and human-aligned behavior while benefiting from the high-level reasoning skills of modern language models.  \\

Our prototype is evaluated within a Unity-based simulation replicating the DJI Tello drone in realistic indoor environments, providing a safe and flexible testbed for complex instruction-following scenarios. The results demonstrate that VLLMs can significantly reduce human workload while maintaining effective and safe navigation performance, opening promising avenues for scalable, user-friendly indoor drone autonomy.  \\

\section{Related Work}
VLN has emerged as a crucial paradigm at the intersection of computer vision, natural language processing, and embodied robotics. Classical works such as Anderson et al.~\cite{anderson2018vision} laid the foundation with the Room-to-Room (R2R) benchmark, providing a photorealistic simulator paired with natural language instructions to train and evaluate navigation agents. Since then, a broad range of improvements has been proposed. Survey studies~\cite{wu2022vlnsurvey} have categorized VLN into goal-oriented and route-oriented tasks, while recent benchmarks also explore multi-turn and interactive dialog-based settings.  \\

Recent efforts have advanced the learning frameworks underlying VLN. Episodic Transformer (E.T.)~\cite{pashevich2021episodic} leveraged transformer encoders to track the entire episode history of visual observations and actions, improving long-horizon compositional tasks such as ALFRED \cite{ALFRED20}. Likewise, FlexVLN~\cite{zhang2025flexvln} integrated hierarchical reasoning and large language models to generalize instructions across different datasets, addressing the critical problem of domain adaptation in VLN tasks. SOAT~\cite{moudgil2021soat} introduced scene-object-aware transformers to better ground both coarse-grained scene references and fine-grained object descriptions from language to vision. VLN-R1~\cite{qi2025vlnr1} showed that reinforcement fine-tuning with large vision-language models can achieve end-to-end embodied reasoning from video streams rather than static graphs, paving the way for continuous environment navigation.  \\

Indoor drone navigation adds further challenges to VLN by requiring 3D path planning, altitude control, and collision avoidance in constrained indoor spaces. UAV-VLN~\cite{saxena2025uavvln} proposed an end-to-end system for drones combining large language models with vision-based perception, highlighting robust spatial reasoning and cross-modal grounding for Unmanned Aerial Vehicles (UAV) control. Similarly, AerialVLN~\cite{liu2023aerialvln} introduced a UAV-specific VLN dataset with a continuous simulator based on city-level photorealistic scenes, allowing the agent to reason about height, dynamic weather, and complex spatial relations in flight. Airbert~\cite{guhur2024airbert} focused on domain-adapted pretraining for VLN by leveraging millions of path-instruction pairs mined from online listings, which improved few-shot performance on indoor settings such as R2R and REVERIE.  \\

In terms of datasets, R2R~\cite{anderson2018vision} remains the gold standard for indoor instruction-based navigation, providing realistic panoramas from the Matterport3D scans. Its successors, RxR~\cite{ku2020rxr}, VLN-CE~\cite{Krantz_2023_CVPR}, and LHPR-VLN~\cite{song2025lhprvln}, have extended the challenge to multilingual instructions, continuous control spaces, and long-horizon multi-stage tasks. These expansions address key limitations of the static graph-based navigation paradigm, making them more representative of real-world deployment. Furthermore, DynamicVLN~\cite{sun2025dynamicvln} proposed a benchmark specifically designed to incorporate dynamic changes, such as moving vehicles and changing weather conditions, providing a testbed for agents to handle temporal interruptions with “temporal stop” actions.  \\

These recent advances have also sparked interest in hybrid or hierarchical models that incorporate chain-of-thought reasoning and memory modules to handle long-term dependencies in complex instruction-following. The Multi-Granularity Dynamic Memory (MGDM) module from Song et al.~\cite{song2025lhprvln} is an example, integrating short-term and long-term reasoning to manage sequential subtasks.  \\

In summary, the current state of the art demonstrates a transition from static, graph-based indoor navigation using ground robots to more flexible, continuous, and dynamic navigation for aerial vehicles, incorporating advanced transformer-based multimodal learning, reinforcement fine-tuning, and large-scale domain-adapted pretraining. Despite significant progress, open challenges remain for seamless sim-to-real transfer, robust obstacle avoidance, and reliable human-centered instruction grounding in dynamic, cluttered indoor environments with drones.  \\
\section{Methodology}
\begin{figure}[!t]
    \centering
    \includegraphics[scale = 0.3]{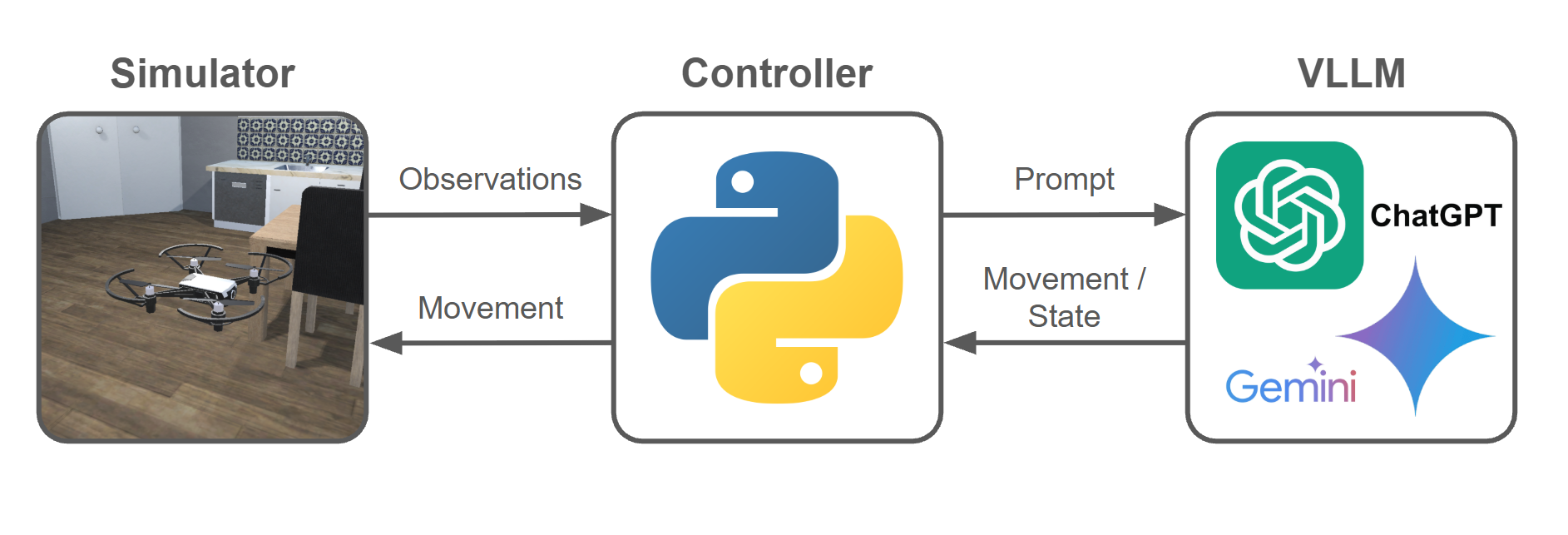}
    \caption{System architecture showing the closed-loop pipeline between the Unity drone simulator, the Python controller, and the VLLM.}
    \label{fig:sistem_arq}
\end{figure}

In this section, we describe the overall architecture and design of the proposed VLN-Pilot system, which combines VLLMs with a rule-based finite-state machine (FSM) to achieve autonomous indoor drone navigation. Our system is designed to operate within a Unity-based realistic simulator of a DJI Tello drone, using the ML-Agents \cite{juliani2020} framework to collect observations and apply actions. The pipeline consists of three main modules. First, the Unity simulator provides visual observations and topological state information through ML-Agents, which are then transmitted to a Python-based controller. Second, the controller forwards these observations to a conversational VLLM (e.g., GPT, Gemini) via a structured prompt, asking it to interpret the current situation and output a high-level movement command as well as the next discrete state of the FSM. Third, the Python controller parses the VLLM's response, sends the corresponding control signals to the simulator to move the drone, and repeats the cycle. A schematic diagram showing the structure of the system is shown in Figure \ref{fig:sistem_arq}.  \\

This closed-loop interaction allows the VLLM to act as a high-level decision-maker, while the FSM handles low-level drone execution. In the following subsections, we explain each of these components in detail.  \\
\subsection{Unity Simulator}
The Unity Simulator provides the virtual environment where the drone agent operates and is tested. Built on the Unity engine, it models realistic indoor scenes with basic textures, lighting, and physical interactions.\\

\begin{figure}[!t]
    \centering
    \includegraphics[scale = 0.3]{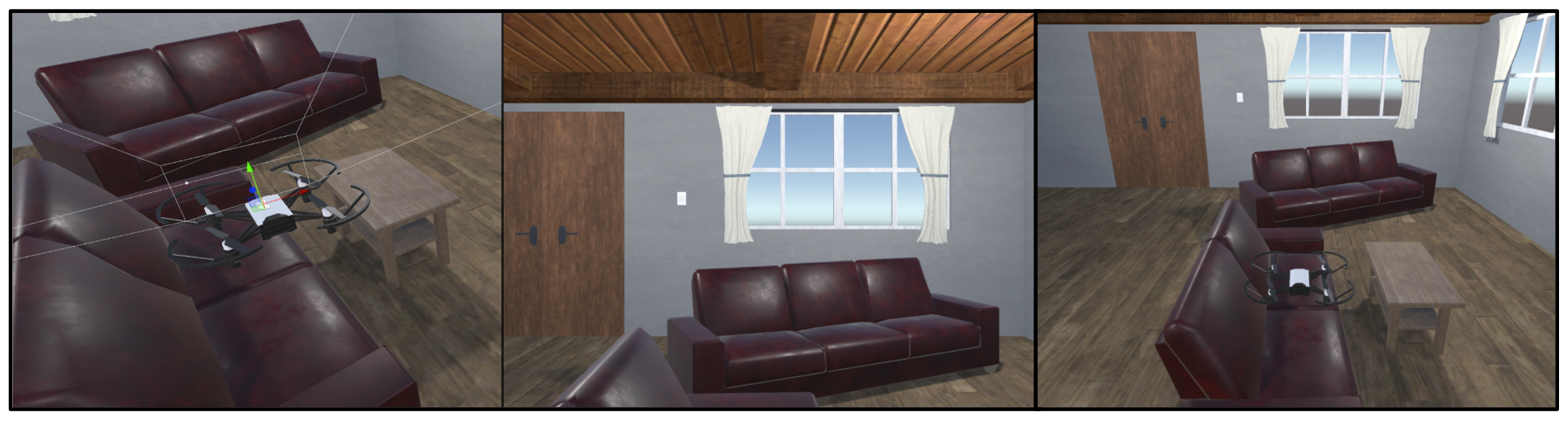}
    \caption{Simulated drone (left) and camera configuration, including front (middle) and rear (right) RGB views. The red, green, and blue arrows correspond to the X, Y, and Z axes, respectively.}
    \label{fig:cameras}
\end{figure}

We simulate the DJI Tello drone, including its camera view, flight dynamics, and collision detection, to approximate real-world conditions. Using the ML-Agents toolkit, the simulator sends the following observations to the controller: a frontal RGB image, a rear RGB image to visualize the drone's behavior, the drone's position in \textit{X}, \textit{Y}, and \textit{Z} coordinates, its orientation around the vertical axis (yaw), and the collision status of the drone. The simulator is in charge of calculating whether the drone has collided with the environment, using Unity's physics engine. As a response from the controller, the drone receives a motion command that specifies its movement along the left–right, front–back, and up–down axes, as well as its rotation to the left or right. Figure~\ref{fig:cameras} shows the drone and camera placements used in the simulation.

\begin{figure}[!t]
    \centering
    \includegraphics[scale = 0.27]{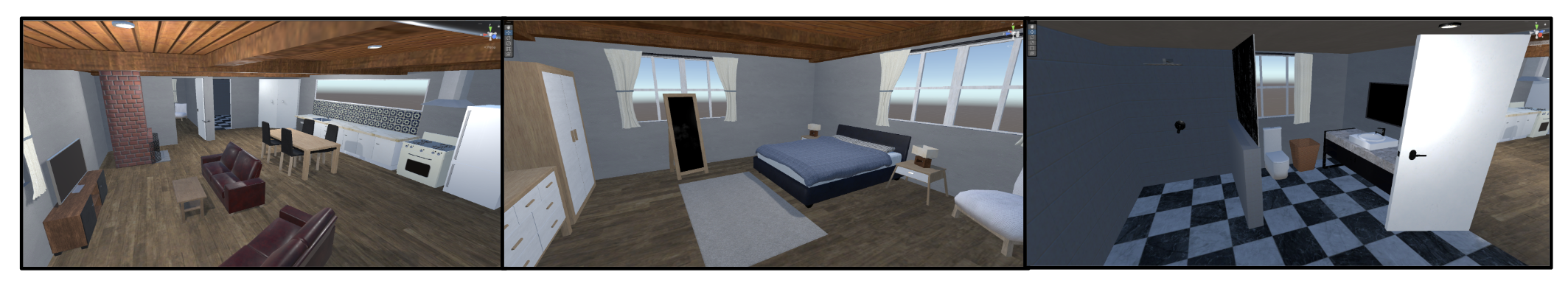}
    \caption{Layout of the simulated furnished cabin environment, with the living room on the left, the bedroom in the center, and the bathroom on the right.}
    \label{fig:rooms_vis}
\end{figure}

The indoor environment used in this work is based on the Furnished Cabin asset by NatureManufacture \cite{unityFurnishedCabin}, available from the Unity Asset Store. The simulated environment consists of a furnished cabin with multiple connected rooms, living room that is also a kitchen, bedroom and bathroom. Figure \ref{fig:rooms_vis} shows the layout of the rooms that make up the simulated environment.  \\

\subsection{Python Controller}
The Python controller is responsible for coordinating the communication between the Unity-based simulator and the VLLM. In addition, the controller is responsible for launching the experiments, controlling the FSM that defines the behavior of the drone in different situations and checking if the target has been reached. The controller consists of several key components described below. \\

The controller manages a finite state machine (FSM) to define the behavior that the drone should follow at each step of the system’s execution. This FSM establishes a sequence of semantic states that guide the drone’s actions throughout the navigation and interaction tasks. These states include room recognition, door navigation, object search, and object description, among others. A detailed description of each state is presented in Table~\ref{tab:states}.

\renewcommand{\arraystretch}{1.2}
\begin{table}[!t]
\centering
\caption{States of the drone navigation and interaction system.}
\begin{tabular}{|l|p{9cm}|}
\hline
\textbf{State} & \textbf{Description} \\
\hline
Start & The initial state where the system is activated. \\
\hline
    Recognize room & The agent tries to determine the room in which it is currently located. \\
\hline
Search open door & If the target room is different from the current one, the system searches for an open door to navigate through. \\
\hline
Orient towards door & Aligns the drone to face the detected open door before crossing it. \\
\hline
Go through door & Commands the drone to move through the recognized and oriented door. \\
\hline
Stay on room & \textit{(Final state)}. When the target room is reached and no object is to be found. \\
\hline
Search object & Once inside the target room, the system searches for the specific object of interest. \\
\hline
Reach object & Guides the drone closer to the object. \\
\hline
Describe object & \textit{(Final state)}. When close to the object, the system describes its position and orientation with respect to the target. \\
\hline
\end{tabular}
\label{tab:states}
\end{table}

\begin{figure}[!htb]
    \centering
    \includegraphics[scale = 0.35]{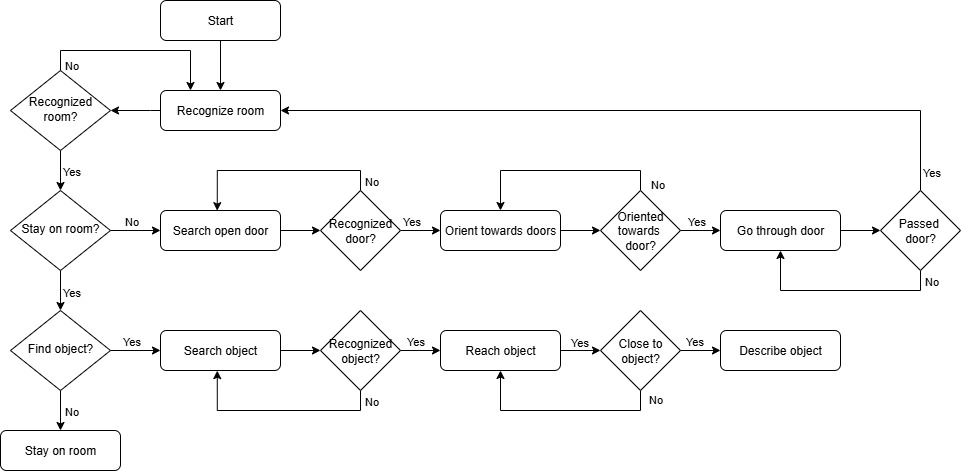}
    \caption{FSM diagram of the drone controller, showing the main execution states (rectangles) and the transition conditions (diamonds) evaluated through queries to the VLLM.}
    \label{fig:states}
\end{figure}

To perform transitions between states, the controller queries the VLLM to determine which should be the next state based on the current visual information, the current state, and the previous movement. Figure \ref{fig:states} illustrates the complete FSM with the conditions and questions asked to the model, used to decide the changes in state.

The system uses a topological map to represent the connectivity of the environment in a simplified way. Instead of relying on precise metric coordinates, the topological map defines a graph where each node corresponds to a room (such as the living room, bedroom, or bathroom), and each edge represents a navigable connection between rooms (bedroom connects to living room). \\

Finally, the controller uses a set of predefined motion commands, each
with a specified duration and displacement. These commands are grouped
by motion type and are labeled for easy reference in the control logic,
as summarized in Table~\ref{tab:simple_motion_commands}.


\begin{table}[!t]
\centering
\caption{Predefined drone motion commands.}
\label{tab:simple_motion_commands}
\begin{tabular}{|c|c|c|}
\hline
\textbf{Command} & \textbf{Type} & \textbf{Details} \\
\hline

A1--A3 & Forward & 10, 25, 50 cm \\
\hline

B1--B3 & Rotate Right & 15°, 45°, 90° \\
\hline

C1--C3 & Rotate Left & 15°, 45°, 90° \\
\hline

D1, D2 & Lateral & 10 cm left, right \\
\hline

E & None & No movement \\
\hline

\end{tabular}
\end{table}

\subsection{VLLM Module}

This module is responsible for interacting with the chosen VLLM by creating appropriate prompts based on the current state of the drone and its observations. \\

The prompt is divided into three parts. First, a general description is provided for all states (see Appendix \ref{app:sys_}). This description explains the expected behaviour of the model, which must act as a drone pilot that navigates safely to reach a specified target. It also clarifies the type of information the model receives as input, and how the response should be structured to ensure effective and coherent control. \\

The second part of the prompt depends on the current state of the system. For each state, the specific goal that the model must achieve is described, along with a set of rules it should follow. Additionally, the prompt lists the allowed movement commands available in that state and specifies the possible next states the model can transition to, depending on the drone's situation. An example of a prompt for one state can be found in Appendix (see Appendix \ref{app:sys_state}).Table \ref{tab:fsm_style} shows detailed information on the generated prompt for each state.\\

\begin{table}[!t]
\caption{Summary of FSM states, allowed movement commands, and possible transitions.}
\label{tab:fsm_style}
\begin{tabular}{|p{0.45\textwidth}|p{0.2\textwidth}|p{0.32\textwidth}|}
\hline
\textbf{State and Goal} & \textbf{Allowed Moves} & \textbf{Next States} \\
\hline

\textbf{Recognize Room} \newline Identify the current room. &
A1, B1--B3, C1--C3 &
Stay on Room, Search Object, Search open Door \\
\hline


\textbf{Stay on Room} \newline Remain stationary in the target room. &
E &
Final \\
\hline

\textbf{Search Object} \newline Locate and center the target object. &
A1, A2, B1--B3, C1--C3 &
Search Object, Reach Object \\
\hline

\textbf{Reach Object} \newline Approach and align with the object. &
A1, B1, C1, D1, D2, E &
Describe Object, Search Object, Reach Object \\
\hline

\textbf{Describe Object} \newline Stop and describe the object. &
E &
Final \\
\hline

\textbf{Search Open Door} \newline Look for an open door to the goal. &
A1, B1--B3, C1--C3, E &
Search open Door, Oriented Towards Door \\
\hline

\textbf{Oriented Towards Door} \newline Align with the open door. &
A1, B1, C1, D1, D2 &
Go Through Door, Search open Door \\
\hline

\textbf{Go Through Door} \newline Enter the next room through the door. &
A1 &
Position in Center of Room, Oriented Towards Door \\
\hline

\end{tabular}
\end{table}

The model receives as input the user's query, the drone's frontal image, the topological map of the house, the previous state, and the previous movement. Based on this information, the model must determine the next movement the drone should execute, update the system's state, identify the current room it believes the drone is in, provide a brief description of the visual input, and explain the reasoning behind its decisions. 

Finally, a last prompt is added to provide information about the specific output format (see Appendix \ref{app:sys_out}). For an VLLM answer example see Appendix \ref{app:output}.\\

\section{Experimental Setup}

\subsection{Simulation}
\begin{figure}[!t]
    \centering
    \begin{subfigure}[b]{0.4\textwidth}
        \centering
        \includegraphics[width=\textwidth]{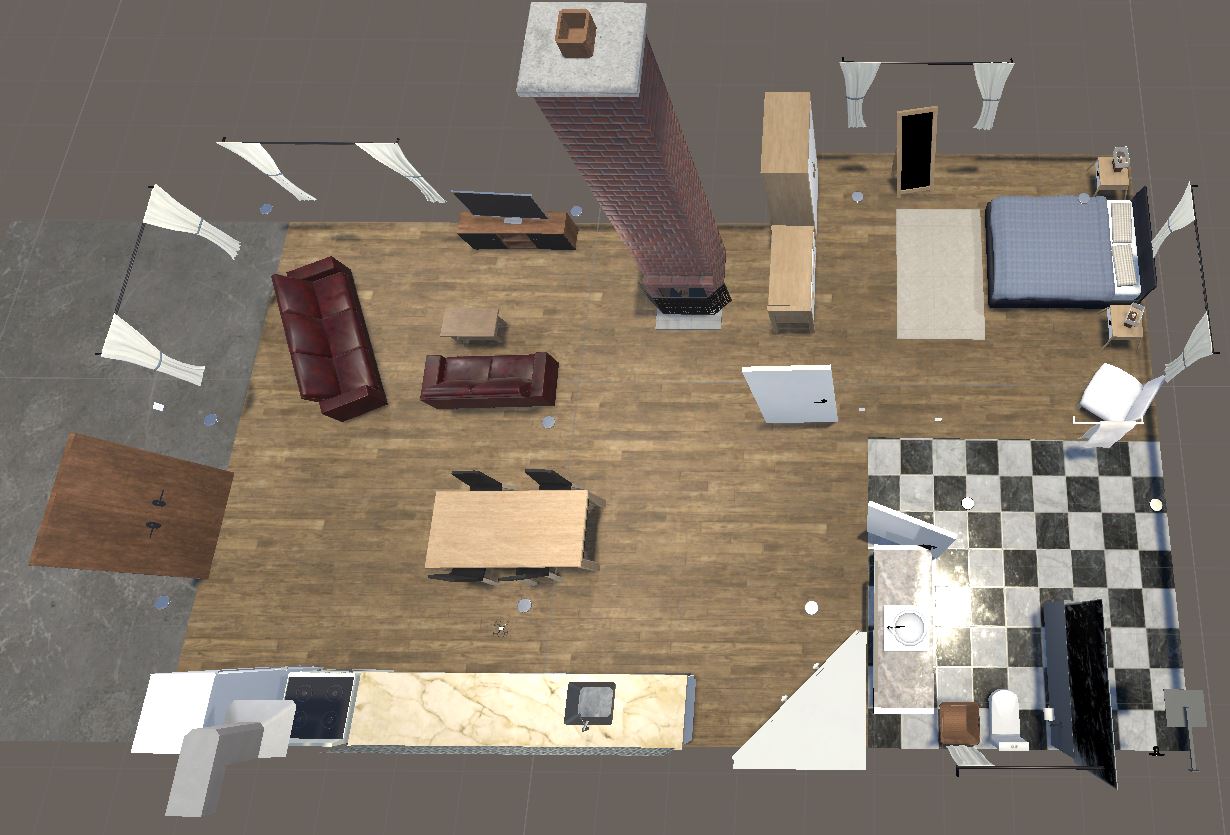}
        \caption{Zenithal view of the environment.}
        \label{fig:rooms_a}
    \end{subfigure}
    \begin{subfigure}[b]{0.41\textwidth}
        \centering
        \includegraphics[width=\textwidth]{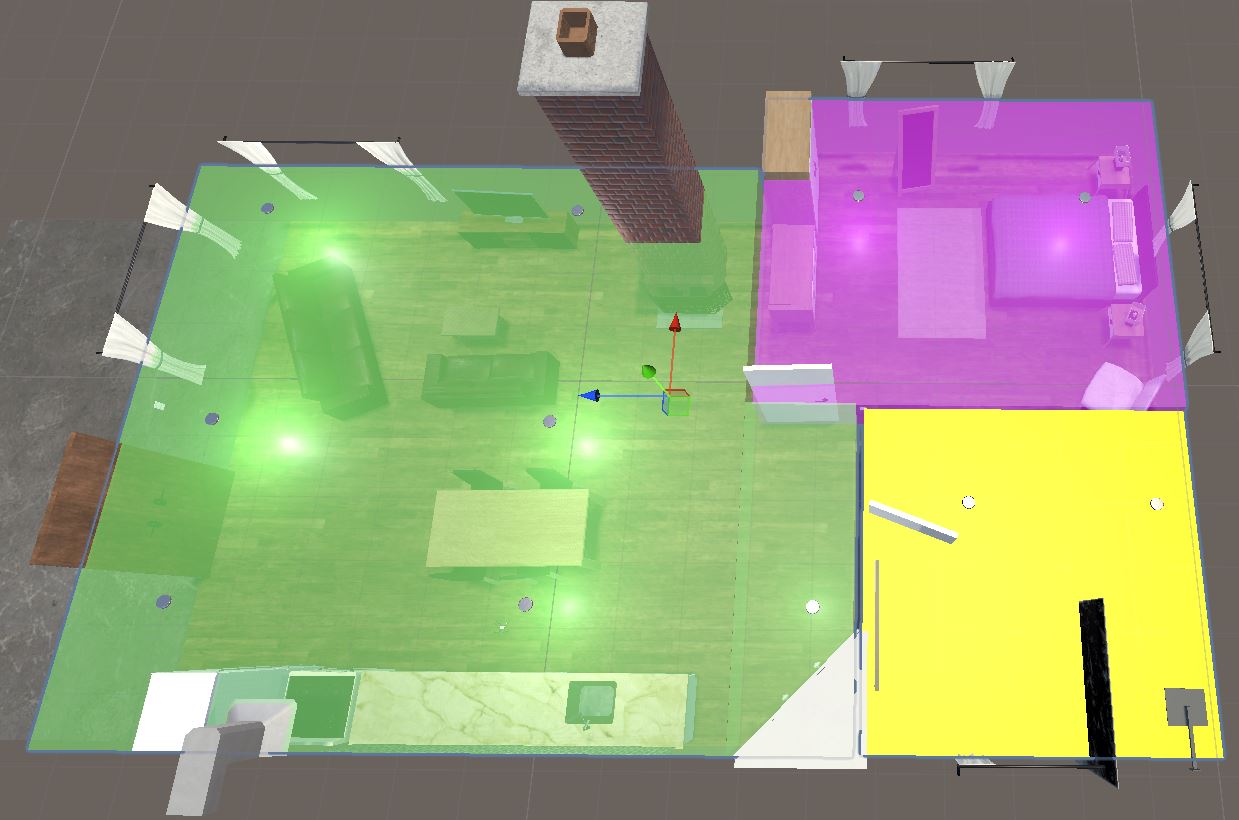}
        \caption{Colored room diagram with labels.}
        \label{fig:rooms_b}
    \end{subfigure}
    \caption{Room diagrams of the simulator. The living room is highlighted in green, the bedroom in magenta, and the bathroom in yellow.}
    \label{fig:rooms}
\end{figure}

The main objectives of the system are to navigate to a specified room and to recognize relevant objects described in the user query. Based on the drone's \textit{X} and \textit{Z} coordinates, Unity simulator non-vertical axes, the system can determine the room in which the drone is currently located. Figure \ref{fig:rooms} shows the room diagram of the simulator. \\

\begin{figure}[!t]
    \centering
    \begin{subfigure}[b]{0.45\textwidth}
        \centering
        \includegraphics[height=3.75cm]{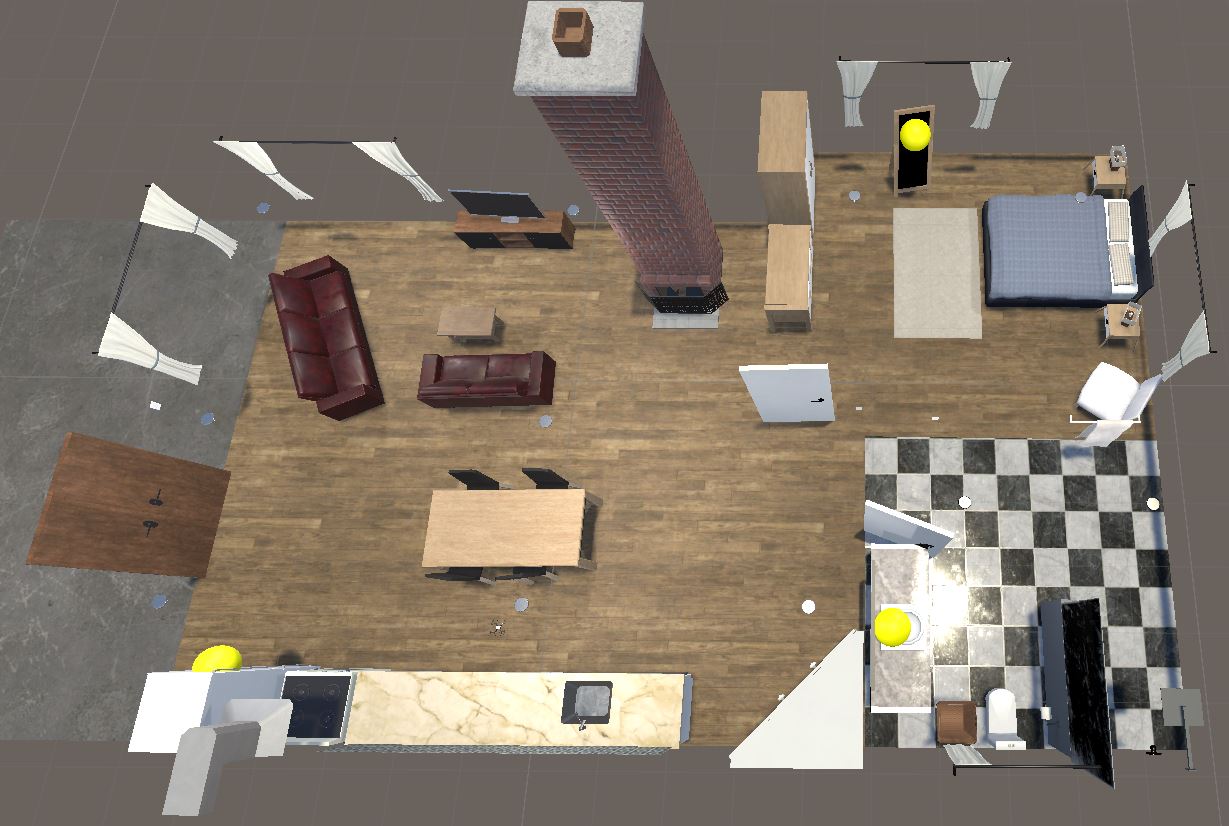}
        \caption{Positions of the navigation targets within the environment.}
        \label{fig:goals_a}
    \end{subfigure}
    \begin{subfigure}[b]{0.45\textwidth}
        \centering
        \includegraphics[width=\textwidth, height=3.75cm]{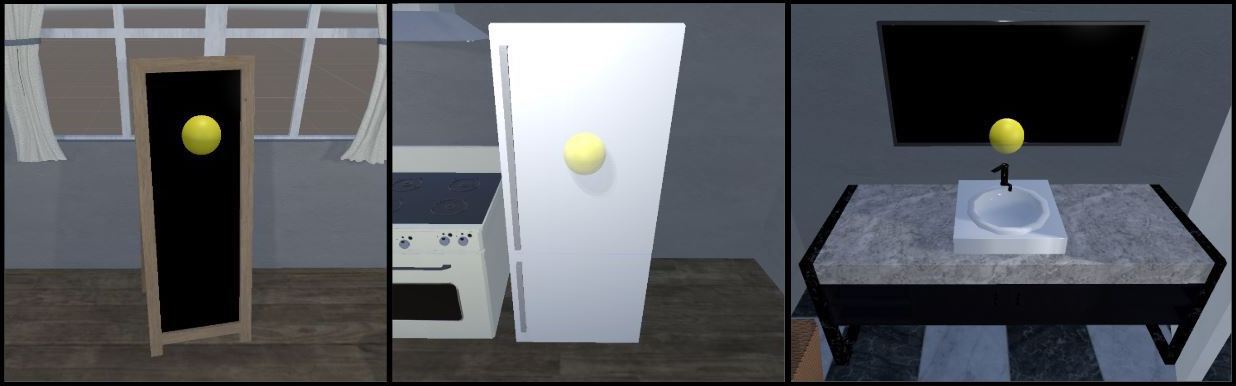}
        \caption{Example of visual views for each target.}
        \label{fig:goals_b}
    \end{subfigure}
    \caption{Objectives: (a) shows the positions of the targets in the environment, while (b) illustrates the camera views associated with each target.}
    \label{fig:goals}
\end{figure}

 The possible targets include a refrigerator located in the living room; a mirror  in the bedroom; and a sink located in the bathroom. Figure~\ref{fig:goals} shows the positions of the targets used in the simulator. \\

\begin{figure}[!htb]
    \centering
    \includegraphics[scale = 0.2]{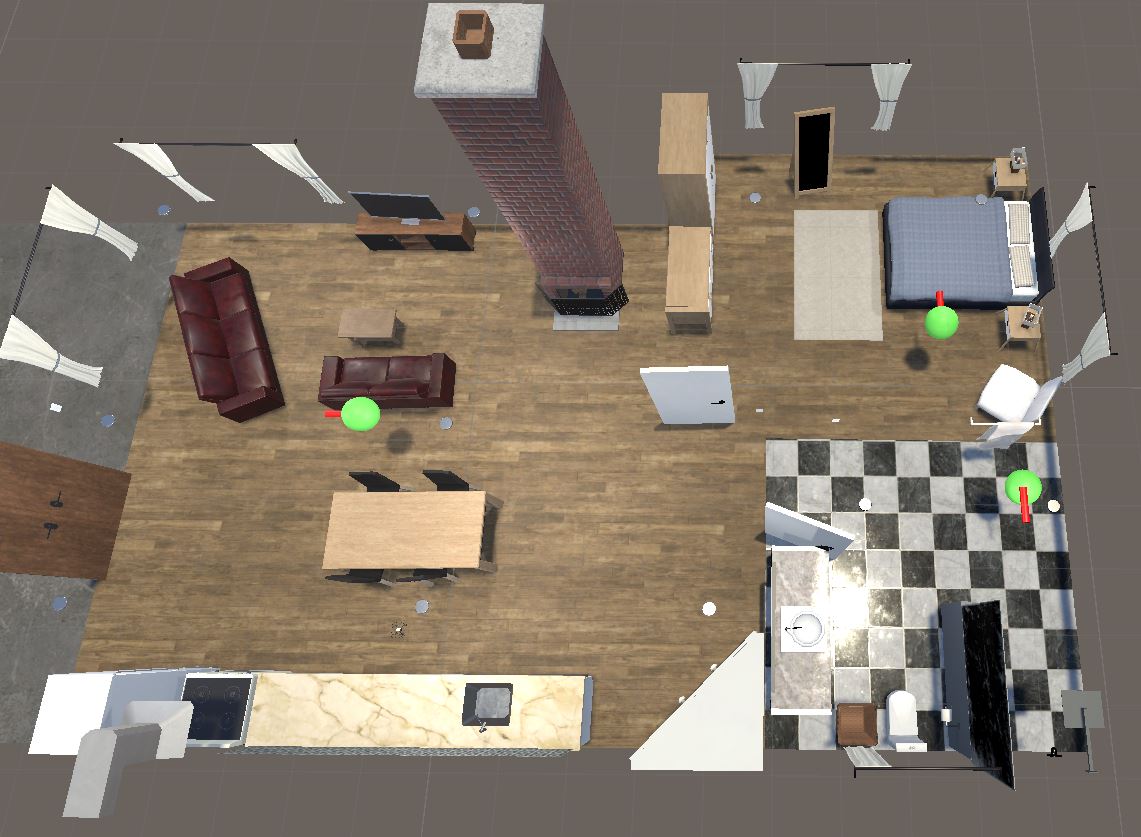}
    \caption{Spawn points in the environment. The green sphere marks the starting position, and the red bar shows the initial orientation.}

    \label{fig:spawns}
\end{figure}

To evaluate the behavior of the navigation system under diverse initial conditions, we define 3 spawn points within the environment. Each spawn point represents a different starting location for the drone, the selected spawn points can be seen in Figure \ref{fig:spawns}.

On the other hand, the Python controller executes the simulation by specifying the drone's initial position, the query that defines the navigation goal, the number of repetitions (5) and the maximum number of allowed execution steps (50). In addition, this controller is also responsible for monitoring whether the query objective has been reached. It determines the room in which the drone is located and verifies when it is close enough and properly oriented toward a possible target. The execution ends under any of the following conditions: the target is successfully reached, the model incorrectly believes it has reached the target, the drone collides with an obstacle or the maximum number of allowed execution steps is exceeded. \\

\subsection{VLLM Server}

To run the experiments, we designed a web server to handle the requests to the VLLMs using the following test setup: an Intel(R) Core(TM) i9-7920X CPU @ 2.90 GHz with 64GB of G.Skill DDR4 RAM (4×16 GB modules), mounted on an ASUS WS X299 SAGE motherboard (X299 chipset). The system was equipped with an NVIDIA GeForce RTX 4090 GPU and ran on Ubuntu 24.04.

For server-side communication, we used the Flask framework. We run experiments using VLLM models: GPT and Gemini. This experiments were performed using the openai Python module and the google-genai library respectively. Specifically, we used "gpt-4.1" and "gemini-2.5-flash" models.
\section{Results and Discussion}

In this section, we describe the experiments conducted to evaluate the performance of the GPT and Gemini VLLMs in the context of our designed task. Table \ref{tab:results} summarizes the outcomes obtained after five executions from the same initial room (spawn point) using the same specific query. The evaluation focuses on metrics such as goal achievement, number of collisions, and whether the agent exceeded the maximum number of allowed steps.

\renewcommand{\arraystretch}{1.2}
\begin{table}[ht]
\centering
\caption{Comparison of GPT and Gemini results grouped by starting room with detailed achievement scores, number of collisions, and maximum steps exceeded.}
\begin{tabular}{|l|p{4.2cm}|ccc|ccc|}
\hline
\textbf{Starting Room} & \textbf{Query} & \multicolumn{3}{c|}{\textbf{GPT}} & \multicolumn{3}{c|}{\textbf{Gemini}} \\
\cline{3-8}
 & & \textbf{Ach.} & \textbf{Coll.} & \textbf{Steps} & \textbf{Ach.} & \textbf{Coll.} & \textbf{Steps} \\
\hline
\multirow{4}{*}{\shortstack[l]{Living Room \\ and Kitchen}} 
 & Go to the bathroom & 2/5 & 3 & -- & 2/5 & 1 & 2 \\
 & Go to the bedroom & 4/5 & -- & 1 & 0/5 & 1 & 4 \\
 & Go to the living room/kitchen & 5/5 & -- & -- & 5/5 & -- & -- \\
 & Find the refrigerator in the living room or kitchen & 4/5 & -- & 1 & 0/5 & -- & 5 \\
\hline
\multirow{3}{*}{Bedroom} 
 & Go to the bedroom & 5/5 & -- & -- & 5/5 & -- & -- \\
 & Go to the living room/kitchen & 4/5 & -- & -- & 2/5 & 3 & -- \\
 & Find the mirror in the bedroom & 4/5 & -- & 1 & 4/5 & -- & 1 \\
\hline
\multirow{3}{*}{Bathroom} 
 & Go to the bathroom & 5/5 & -- & -- & 5/5 & -- & -- \\
 & Go to the living room/kitchen & 5/5 & -- & -- & 5/5 & -- & -- \\
 & Find the sink in the bathroom & 5/5 & -- & -- & 5/5 & -- & -- \\
\hline
\end{tabular}
\label{tab:results}
\end{table}

\begin{figure}[!htb]
    \centering
    \includegraphics[scale = 0.3]{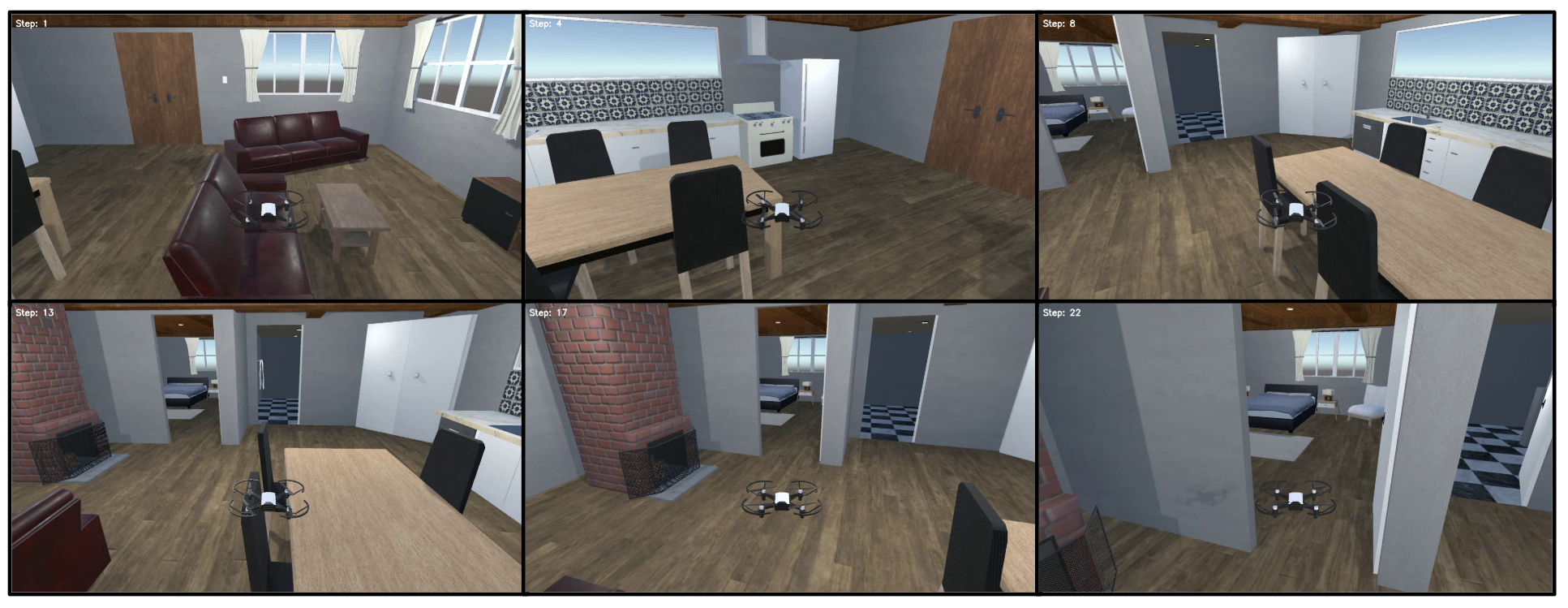}
    \caption{Example of correct result achieved by GPT model.}
    \label{fig:ex_correct}
\end{figure}

\begin{figure}[!htb]
    \centering
    \includegraphics[scale = 0.3]{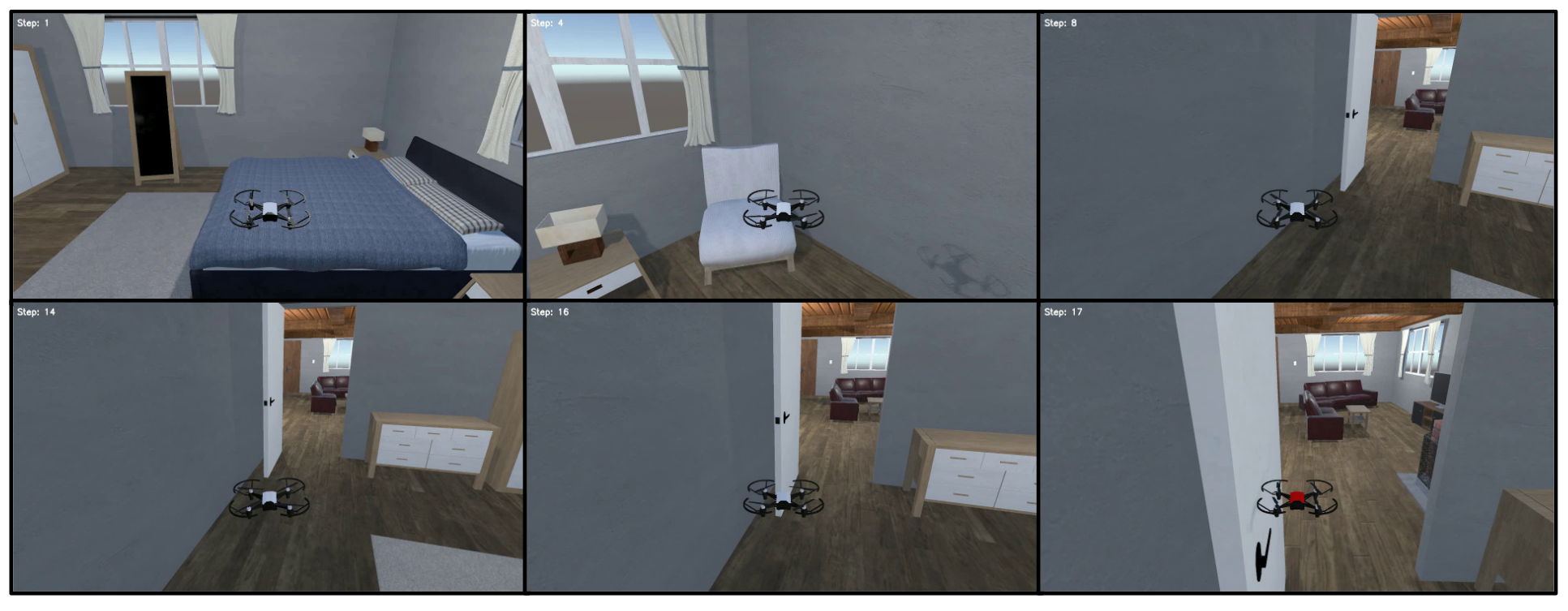}
    \caption{Example of incorrect result with a collision generated by Gemini model.}
    \label{fig:ex_incorrect}
\end{figure}\textbf{}

From Table \ref{tab:results}, the GPT model outperformed the Gemini model in different cases, such as the queries "Go to the bedroom" and "Find the refrigerator", both starting from the living room/kitchen area. Also in the query "Go to the living room/kitchen from the starting spawn bedroom. In these scenarios, Gemini model reached the maximum number of allowed steps more frequently and exhibited a higher number of collision failures, generating less efficient navigation.

GPT demonstrated promising capabilities in our simulations, effectively leveraging the semantic graph and state transitions to guide the drone toward target rooms. In many test cases, GPT was able to interpret the room connectivity graph correctly, selecting the appropriate doorway to cross and maintaining consistent orientation strategies to approach the target. In particular, GPT exhibited a tendency to interpret “being near the door” as requiring the drone to be almost directly adjacent to the doorway before executing the crossing maneuver, leading to high success rates in recognizing and confirming the target room upon entry. This behavior aligned well with the proximity thresholds defined in the simulation environment, facilitating clear decisions for crossing transitions and contributing to reliable target room identification.

To illustrate the experimental results, Figure \ref{fig:ex_correct} presents a successful execution example using GPT, where the drone starts in the living room/kitchen area with the goal of reaching the bedroom. It performs a search to identify open doors, finds two possible options, and decides to proceed toward the door whose visible contents appear to match the target room. Finally, it aligns itself with the doorway and crosses through it.

On the other side, Gemini exhibited a different interpretation of spatial relationships compared to GPT. It perceived “being near the door” as maintaining a larger distance from the doorway and applied a strict criterion for doorway centering, requiring the doorway to appear precisely centered within the drone’s field of view before proceeding to cross. This behavior led Gemini to continuously attempt fine alignment adjustments from relatively distant positions, resulting in oscillatory re-centering behaviors that frequently prevented the drone from approaching the door closely enough to cross successfully.

To address this oscillation loop, the prompt provided to Gemini was modified to explicitly instruct the drone to approach the doorway more closely before executing alignment adjustments. While this modification reduced the amplitude of the oscillatory re-alignment, it introduced a new challenge: the drone, following the instruction to approach more aggressively, entered configurations where it collided with objects near the doorway, including door frames and nearby obstacles, due to insufficient collision avoidance behaviour during close-quarters maneuvering. This highlights the fragility of direct prompt-based adjustments in VLLM-guided navigation, where resolving one limitation (oscillation) can inadvertently expose new issues (collisions) when spatial reasoning remains incomplete.

An additional limitation identified during experimentation was that the concept of "centered" in VLLMs lacks the necessary spatial awareness to determine whether the drone can actually pass through the door based on its physical dimensions. In several cases, this led the drone to collide with door frames, with minor contacts occurring on the tips of the drone’s propellers as it attempted to cross while believing it was sufficiently centered. This indicates that VLLMs operate under a simplified notion of spatial alignment, lacking true volumetric awareness of the drone’s bounding dimensions relative to the available doorway clearance.

Figure \ref{fig:ex_incorrect} shows an unsuccessful execution example using Gemini, where the drone ends up colliding with its left propeller because the VLLM is not aware of the dimensions of the drone it is guiding during navigation.

\section{Conclusion and Future Work}

In this study, we evaluated the use of GPT and Gemini to guide a simulated drone toward room-level goals and object localization using LLMs informed by semantic graphs of room connectivity and state graphs for decision-making support.

The findings suggest that GPT currently provides a more practical baseline for LLM-driven drone guidance in structured indoor environments, benefiting from its tighter proximity thresholds and willingness to commit to crossing actions when conditions are met adequately. Meanwhile, Gemini’s cautious approach and strict spatial alignment criteria, while potentially useful in scenarios requiring precise alignment, may require parameter adjustments or prompt engineering interventions to improve its effectiveness in navigation tasks requiring doorways traversal.

Additionally, there are key differences and shared limitations between both models. One notable difference lies in the spatial interpretation strategies of the models: GPT’s preference for tight proximity thresholds enabled the drone to cross doorways consistently and confidently, while Gemini’s approach, which required maintaining a greater distance from the door while ensuring precise centering, often required explicit modifications to the prompt to encourage it to approach the door more closely.

In terms of decision stability, GPT demonstrated a stable and direct approach when executing crossing maneuvers, whereas Gemini’s strict adherence to perfect centering led to repeated, minor alignment adjustments that resulted in oscillatory behaviors, preventing effective forward progress. Attempts to mitigate this oscillation through prompt modifications did reduce the re-alignment loops; however, they introduced new challenges, as the drone, following the revised instructions, began to collide with nearby elements around doorways. This behavior underscores the sensitivity of Gemini’s navigation performance to prompt instructions and highlights the limitations of relying solely on textual adjustments without incorporating structured spatial constraints.

Furthermore, a shared limitation across both models became apparent in their inability to reason about the drone’s physical dimensions relative to the available doorway clearance. Without awareness of the drone’s bounding box, both GPT and Gemini occasionally guided the drone into situations where it collided with door frames or clipped its propellers, despite the models perceiving the drone as correctly aligned for crossing. This limitation emphasizes the need for the integration of volumetric awareness within LLM-driven navigation frameworks to enhance the safety and reliability of drone maneuvering in constrained indoor environments.

Future work will explore fine-tuning the prompt structures and adjusting the semantic-state integration to harmonize Gemini’s cautious strategies with practical navigation needs, while further investigating how these LLMs interpret spatial constraints when grounded in real-world drone perception pipelines.

\bibliographystyle{splncs04}
\bibliography{bibliography}

\begin{thebibliography}{10}
\providecommand{\url}[1]{\texttt{#1}}
\providecommand{\urlprefix}{URL }
\providecommand{\doi}[1]{https://doi.org/#1}

\bibitem{anderson2018vision}
Anderson, P., Wu, Q., Teney, D., Bruce, J., Johnson, M., Gould, S., van~den Hengel, A.: Vision-and-language navigation: Interpreting visually-grounded navigation instructions in real environments. In: CVPR (2018)

\bibitem{guhur2024airbert}
Guhur, P.L., Tapaswi, M., Chen, S., Laptev, I., Schmid, C.: Airbert: In-domain pretraining for vision-and-language navigation. arXiv preprint arXiv:2402.01418  (2024)

\bibitem{juliani2020}
Juliani, A., Berges, V.P., Teng, E., Cohen, A., Harper, J., Elion, C., Goy, C., Gao, Y., Henry, H., Mattar, M., Lange, D.: Unity: A general platform for intelligent agents. arXiv preprint arXiv:1809.02627  (2020), \url{https://arxiv.org/pdf/1809.02627.pdf}

\bibitem{Krantz_2023_CVPR}
Krantz, J., Banerjee, S., Zhu, W., Corso, J., Anderson, P., Lee, S., Thomason, J.: Iterative vision-and-language navigation. In: Proceedings of the IEEE/CVF Conference on Computer Vision and Pattern Recognition (CVPR). pp. 14921--14930 (June 2023)

\bibitem{ku2020rxr}
Ku, A., Anderson, P., Patel, R., Ie, E., Baldridge, J.: Room-across-room: Multilingual vision-and-language navigation with dense spatiotemporal grounding. In: EMNLP (2020), \url{https://arxiv.org/abs/2010.07954}

\bibitem{liu2023aerialvln}
Liu, S., Zhang, H., Qi, Y., Wang, P., Zhang, Y., Wu, Q.: Aerialvln: Vision-and-language navigation for uavs. In: ICCV (2023)

\bibitem{moudgil2021soat}
Moudgil, A., Majumdar, A., Agrawal, H., Lee, S., Batra, D.: Soat: A scene- and object-aware transformer for vision-and-language navigation. In: NeurIPS (2021)

\bibitem{unityFurnishedCabin}
NatureManufacture: Furnished cabin (2017), \url{https://assetstore.unity.com/packages/3d/environments/urban/furnished-cabin-71426}, unity Asset Store

\bibitem{pashevich2021episodic}
Pashevich, A., Schmid, C., Sun, C.: Episodic transformer for vision-and-language navigation. In: NeurIPS (2021)

\bibitem{qi2025vlnr1}
Qi, Z., Zhang, Z., Yu, Y., Wang, J., Zhao, H.: Vln-r1: Vision-language navigation via reinforcement fine-tuning. arXiv preprint arXiv:2506.17221  (2025)

\bibitem{saxena2025uavvln}
Saxena, P., Raghuvanshi, N., Goveas, N.: Uav-vln: End-to-end vision language guided navigation for uavs. arXiv preprint arXiv:2504.21432  (2025)

\bibitem{ALFRED20}
Shridhar, M., Thomason, J., Gordon, D., Bisk, Y., Han, W., Mottaghi, R., Zettlemoyer, L., Fox, D.: {ALFRED: A Benchmark for Interpreting Grounded Instructions for Everyday Tasks}. In: The IEEE Conference on Computer Vision and Pattern Recognition (CVPR) (2020), \url{https://arxiv.org/abs/1912.01734}

\bibitem{song2025lhprvln}
Song, X., Chen, W., Liu, Y., Chen, W., Li, G., Lin, L.: Towards long-horizon vision-language navigation: Platform, benchmark and method. In: CVPR (2025)

\bibitem{sun2025dynamicvln}
Sun, Y., Qiu, Y., Aoki, Y.: Dynamicvln: Incorporating dynamics into vision-and-language navigation scenarios. Sensors  (2025)

\bibitem{wu2022vlnsurvey}
Wu, W., Chang, T., Li, X., Yin, Q., Zhu, J.: Vision-language navigation: A survey and taxonomy. IEEE Transactions on Neural Networks and Learning Systems  (2022)

\bibitem{zhang2025flexvln}
Zhang, S., Qiao, Y., Wang, Q., Guo, L., Wei, Z., Liu, J.: Flexvln: Flexible adaptation for diverse vision-and-language navigation tasks. arXiv preprint arXiv:2503.13966  (2025)

\end{thebibliography}

\appendix

\section{Robot system prompt}
\label{app:sys_}

\begin{lstlisting}[language=Python, basicstyle=\ttfamily\small, breaklines=true]
You are an intelligent planning assistant that controls a drone-like robot operating in an indoor environment using visual input.

Your task is to reason over egocentric visual data and contextual information to select the **next atomic movement** and update the robot's internal state.

You will receive the following inputs:

1. A base64-encoded image representing the robot's current camera view.
2. A user query describing the objective (e.g., "Find the red cup", "Go to the kitchen").
3. A topological map in JSON format describing room connectivity and possible transitions.
4. The current finite state machine (FSM) state.
5. The previous movement command the robot executed.

Your output should:

- Select a **single valid movement command** from the allowed list.
- Predict the **next FSM state** according to the robot's behavior and context.
- Identify the **room the robot is currently in**, or return "unknown" if it cannot be determined.
- Describe the **position of any visible door** (left, center, right, not_visible).
- Include a **short visual description** of the current scene.

You will be guided by specific state-based rules and movement constraints 
Never include multiple actions or narrative text.
\end{lstlisting}

\section{State prompt: Search Object}
\label{app:sys_state}

\begin{lstlisting}[language=Python, basicstyle=\ttfamily\small, breaklines=true]

## State: **Search Object**

---

## GOAL
Scan the current room to locate and center the desired object in the field of view.

---

## POLICY RULES

- Use short forward movements or in-place rotations to explore the surroundings.
- Prefer rotating in the same direction as the previous rotation if applicable (e.g., if `previous_movement` was `B2`, use `B1`, `B2`, or `B3`).
- Alternate between rotating and moving forward if the object has not been seen in the last 3 steps.
- If the object is partially visible but not centered, rotate slightly to center it.

---

## MOVEMENT COMMANDS

Only return **ONE** movement command from the list below:

### A. Forward movement:
- `A1`: move forward 10 cm
- `A2`: move forward 25 cm

### B. Left rotation:
- `B1`: rotate 15$\circ$
- `B2`: rotate 45$\circ$
- `B3`: rotate 90$\circ$

### C. Right rotation:
- `C1`: rotate 15$\circ$
- `C2`: rotate 45$\circ$
- `C3`: rotate 90$\circ$

---

## OUTPUT STATE RULES

Choose the next FSM state based on what the robot sees:

- `Search Object`: if the target object is not yet visible or not centered in the frame.
- `Reach Object`: if the object is clearly visible and approximately centered in the image (i.e., within +-10 \% of the center).

---

## NOTES

- Avoid unnecessary movement: do not move forward if the object might be occluded or outside the frame.

\end{lstlisting}

\section{Output prompt}
\label{app:sys_out}

\begin{lstlisting}[language=Python, basicstyle=\ttfamily\small, breaklines=true]

## Output Format

You must respond with a **single valid JSON object** following the exact structure below:

```json
{
  "room": "estimated_room_name_or_unknown",
  "movement": "valid_single_movement_command",
  "state": "next_FSM_state",
  "description": "concise description of the visible environment",
  "door_position": "left | center | right | not_visible"
}
```

### Field Definitions

- `"room"`: Name of the current room (from the topological map), or `"unknown"` if not confidently identified.
- `"movement"`: A single valid movement command (`A1`, `B2`, etc.) allowed by the current FSM state.
- `"state"`: The predicted FSM state after this movement (e.g., `"Search Object"`, `"Reach Object"`).
- `"description"`: A brief and neutral description of what the robot sees and the reasoning on the instructions.
- `"door_position"`: Where the door appears in the frame. Valid values: `"left"`, `"center"`, `"right"`, `"not_visible"`.

### Important

- Do **not** add extra text or explanations outside the JSON block.
- Do **not** return multiple movement commands.
- Always follow the allowed command list and transition logic defined in the state prompt.

Use all available context (image, state, previous movement, user query, and map) to ensure safe and meaningful output.

\end{lstlisting}

\section{VLLM output example}
\label{app:output}

\begin{lstlisting}[language=Python, basicstyle=\ttfamily\small, breaklines=true]

{'room': 'bedroom', 'movement': 'E', 'state': 'Final', 'description': 'A large mirror is visible on the right side of the image, reflecting light. It is positioned near a window and next to a wardrobe. The mirror is mostly in frame and upright.', 'door_position': 'not_visible'}

\end{lstlisting}

\end{document}